\documentclass[10pt,twocolumn,letterpaper]{article}

\usepackage[pagenumbers]{cvpr}

\usepackage{bm}
\usepackage[most]{tcolorbox}
\usepackage{multirow}
\usepackage{colortbl}
\usepackage{arydshln}  
\usepackage{threeparttable}  

\newcommand{\finding}[2]{
    \begin{tcolorbox}[
        colback=white!90!gray,     
        colframe=teal!60!black,     
        arc=5pt,                    
        boxsep=5pt,                 
        left=10pt,                  
        right=10pt,                 
        top=2pt,                    
        bottom=2pt,                 
        boxrule=0.8pt,              
        drop shadow=gray!50!white,  
        enhanced jigsaw             
    ]
    \vspace{-0.1cm}
        \paragraph{\textbf{\textit{}}} #2
    \end{tcolorbox}
    \vspace{-0.1cm}
}
\newcommand{\diff}[1]{\scriptsize\ \textcolor{ForestGreen}{(#1)}}

\definecolor{cvprblue}{rgb}{0.21,0.49,0.74}
\usepackage[pagebackref,breaklinks,colorlinks,allcolors=cvprblue]{hyperref}

\title{Toward Diffusible High-Dimensional Latent Spaces: A Frequency Perspective}

\author{Bolin Lai$^{1,2\,\dagger}$ \quad
XuDong Wang$^{1}$ \quad
Saketh Rambhatla$^{1}$ \quad
James M. Rehg$^{3}$ \quad \\
Zsolt Kira$^{2}$ \quad
Rohit Girdhar$^{1}$ \quad
Ishan Misra$^{1}$ \\
$^1$Meta AI \quad $^2$Georgia Institute of Technology \quad $^3$University of Illinois Urbana-Champaign \\
{\tt\small {bolin.lai,zkira}@gatech.edu \ \{xudongw,rssaketh,rgirdhar,imisra\}@meta.com \ \ jrehg@illinois.edu} \\
{\small \textbf{Project Page: \textcolor{NavyBlue}{\url{https://bolinlai.github.io/projects/FreqWarm}}}}
}

\begin{document}
\maketitle

\let\thefootnote\relax\footnotetext{$^\dagger$This work was done during Bolin's internship at Meta.}

\begin{abstract}
Latent diffusion has become the default paradigm for visual generation, yet we observe a persistent reconstruction–generation trade-off as latent dimensionality increases: higher-capacity autoencoders improve reconstruction fidelity but generation quality eventually declines. We trace this gap to the different behaviors in high-frequency encoding and decoding. Through controlled perturbations in both RGB and latent domains, we analyze encoder/decoder behaviors and find that decoders depend strongly on high-frequency latent components to recover details, whereas encoders under-represent high-frequency contents, yielding insufficient exposure and underfitting in high-frequency bands for diffusion model training. To address this issue, we introduce FreqWarm, a plug-and-play frequency warm-up curriculum that increases early-stage exposure to high-frequency latent signals during diffusion or flow-matching training -- without modifying or retraining the autoencoder. Applied across several high-dimensional autoencoders, FreqWarm consistently improves generation quality: decreasing gFID by 14.11 on Wan2.2-VAE, 6.13 on LTX-VAE, and 4.42 on DC-AE-f32, while remaining architecture-agnostic and compatible with diverse backbones. Our study shows that explicitly managing frequency exposure can successfully turn high-dimensional latent spaces into more diffusible targets.
\end{abstract}    
\section{Introduction}
\label{sec:intro}

Diffusion models have dominated the field of image and video generation in recent years \cite{yang2025cogvideox,hong2023cogvideo,hacohen2024ltx,ma2024sit,peebles2023scalable,bao2023all,rombach2022high,he2022latent}. Early models fit the raw pixels directly, which suffer from the complex distribution in RGB space. Since the emergence of latent diffusion models \cite{rombach2022high}, modeling data distributions in a latent space has become a canonical paradigm because of the reduced dimensional complexity and smooth latent distributions. Hence, the diffusibility of latent spaces determined by autoencoders becomes a key factor for diffusion model performance.

\begin{figure}
  \centering
  \includegraphics[width=\linewidth]{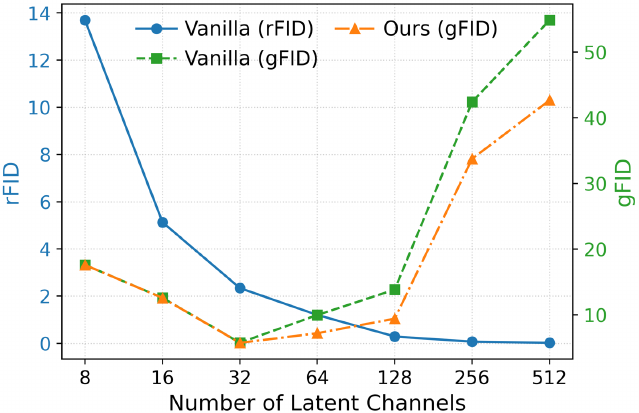}
  \caption{Trade-off between reconstruction and generation. Reconstruction is evaluated by FID between input images and reconstructed images (\ie, rFID). Generation is evaluated by the FID between synthetic images and real images (\ie, gFID). Lower rFID and gFID indicate the better performance. The spatial compression ratio remains 32 for all experiments.
  }
  \label{fig:tradeoff}
\end{figure}

To make the latent space easier for diffusion models to fit (\ie, more diffusible \cite{skorokhodov2025improving}), many efforts have been made to train a stronger autoencoder. Early work uses SD-VAE \cite{rombach2022high} with a spatial compression ratio of 8 and 4 latent channels. To reduce the amount of tokens for computing efficiency, subsequent models further compress the input with higher ratios and compensate the capacity by expanding the number of channels \cite{chen2025deep,labs2025flux1kontextflowmatching,wan2025wan}. However, there is a natural trade-off between \textit{reconstruction} (how well the decoder can recover encoded images) and \textit{generation} (how close the synthetic image distribution is to real image distribution) with regard to the dimension of latent space, as illustrated in \cref{fig:tradeoff}. When the number of channel expands, the reconstruction performance consistently improves (\textcolor{NavyBlue}{blue}), while the generation performance improves at the beginning and then decreases (\textcolor{ForestGreen}{green}). We argue that there are two factors affecting gFID -- the reconstruction fidelity (determined by autoencoders) and the quality of latent embeddings (synthesized by diffusion models). With high dimensionality, the reconstruction fidelity keeps benefiting from the high capacity but overall generation performance eventually declines, which implies a significant drop in latent embedding synthesis. Previous work tends to use a latent space with low dimensions (\eg, 4-channel or 32-channel). How to improve diffusibility in high-dimensional latent space remains understudied, which hinders encoding with higher compression ratios. This is exactly the focus of our paper.

To obtain a diffusible high-dimensional latent space, it is necessary to investigate existing autoencoders to figure out the reason for the reconstruction-generation trade-off. Recent work improves diffusibility by aligning the latent space with semantic embeddings \cite{chen2025softvq,yao2025reconstruction,xu2025exploring}, using hierarchical tokenization \cite{chen2025hieratok,zhang2025diffusion,liu2025hi}, or representing images by 1D seqeunces \cite{yu2024image}. However, most prior studies are motivated intuitively without detailed analysis. Recently, Skorokhodov \etal \cite{skorokhodov2025improving} investigates the change of latent frequency energy with regard to different latent channels, introducing a new perspective to understand the encoding mechanism. 

Inspired by their method, in this paper we conduct a thorough study of how the encoder and decoder react to signals in different frequency bands, resulting in a number of actionable findings. We make frequency perturbation in the latent space and decode high-frequency and low-frequency embeddings separately. Likewise, we also separate frequency bands in the RGB space and foward them to a pre-trained encoder. In our experiments, we find a significant difference in encoding and decoding behaviors, especially for high-frequency signals. Specifically, the decoder greatly relies on high-frequency components in the latent embeddings to reconstruct details, which suggests the importance of synthesizing high-quality embeddings in high-frequency bands. However, we also find it very challenging for the encoder to encode high-frequency information. A portion of extremely high-frequency RGB signals even impede the encoding process, which leads to a lower energy in high-frequency bands of the latent space. The diffusion models thus fail to fit the distribution in the high-frequency band due to under-representation during training. This phenomenon is more prominent in high-dimensional space, thus leading to the trade-off shown in \cref{fig:tradeoff}.

To mitigate this issue, we propose an easy-to-implement \textbf{freq}uency \textbf{warm}-up strategy, termed \textbf{FreqWarm}, which exposes diffusion models to more high-frequency latent embeddings in the early training stage. In contrast to the previous work that focuses on model architectures and training losses, our method does not require any training for autoencoders. Thus we can fully leverage the off-the-shelf autoencoders as the starting point and further improve diffusibility on top of it. We implement our methods on top of a variety of high-dimensional autoencoders. Experiments show that FreqWarm consistently improves generation performance of Wan2.2-VAE by 14.11, LTX-VAE by 6.14 and DC-AE-f32 by 4.42 in gFID. Our method can also generalize to diverse diffusion and flow matching architectures.

Overall, our contributions can be summarized as follows:

\begin{itemize}
    \item We conduct the first analysis to reveal the different behaviors of encoders and decoders to signals with different frequencies. 

    \item Inspired by our findings, we propose a plug-and-play method to warm up latent spaces in frequency domain to improve the diffusibility for diffusion and flow matching model training, without re-training the autoencoders.

    \item Extensive experiments suggest that our method consistently improves the generation performance for high-dimensional latent spaces defined by various autoencoders, providing a new training recipe for future work.
\end{itemize}

\section{Related Work}
\label{sec:related_work}

\noindent\textbf{Autoencoders for Visual Generation}\quad Modern diffusion models depend on diffusible latent spaces for generative training. Many studies have been conducted to improve diffusibility by training a better autoencoder \cite{kouzelis2025eq,zhang2025gpstoken,lu2025atoken,sargent2025flow,lee2025latent,qiu2025image,vallaeys2025ssdd,wu2025h3ae,medi2025missing,mahapatra2025progressive}. Early work uses SD-VAE \cite{rombach2022high} to tokenize RGB images into 4-channel embeddings with a spatial compression ratio of 8. Recent work pursues higher compression ratios coupled with more latent channels \cite{wan2025wan,hacohen2024ltx,chen2025dc,yu2024image}. DC-AE \cite{chen2025deep} recently achieves compression ratios of 32 and 64, reducing the computation complexity in diffusion models. To incorporate more semantics in latent spaces, SoftVQ-VAE \cite{chen2025softvq} and VideoREPA \cite{zhang2025videorepa} are proposed to align the latent representations with visual features obtained from visual models. Recent studies find that visual foundation models can be directly used as encoders by training a paired decoder \cite{shi2025latent,zheng2025diffusion,bi2025vision}. VA-VAE \cite{yao2025reconstruction} and ReaLS \cite{xu2025exploring} use features from frozen visual foundation models (such as DINO \cite{caron2021emerging}, SAM \cite{kirillov2023segment} and MAE \cite{he2022masked}) as guidance for alignment. Besides, many investigations also focus on improving tokenization and detokenization strategies, including hierarchical encoding \cite{chen2025hieratok,liu2025hi,zhang2025diffusion} and multi-step decoding \cite{zhao2025epsilon}. Most of prior work is driven by intuitive motivation and lacks detailed analysis on latent spaces. SE-VAE \cite{skorokhodov2025improving} is the most relevant study to our work. They analyze frequency distributions in latent spaces, which introduce a new tool to interpret the mechanism of autoencoders. Inspired by this work, we further conduct a detailed analysis on high-dimensional autoencoders to understand the reaction to frequency perturbations. Motivated by our findings, we propose a plug-and-play frequency warm-up recipe 
that can be used jointly with previous novel architecture designs 
to improve diffusibility without re-training autoencoders.

\vspace{0.1cm}
\noindent\textbf{Diffusion-based Generation}\quad Early diffusion models corrupt input images by adding noise iteratively and adapt the UNet structure for noise estimation \cite{rombach2022high,he2022latent}, establishing today’s default training recipe. Transformer backbones subsequently replaced U-Nets \cite{wan2025wan,kong2024hunyuanvideo,yang2025cogvideox}. U-ViT \cite{bao2023all} shows that plain ViTs can serve as effective diffusion backbones, and DiT \cite{peebles2023scalable} scales this paradigm to state-of-the-art ImageNet synthesis in latent space. Building on DiT, SiT \cite{ma2024sit} unifies diffusion and flow-style training under an interpolant framework and reports consistent gains at fixed compute, reinforcing transformers as the base model family for modern diffusion. For video synthesis, foundational works adapt image diffusion to the temporal domain and introduce cascaded pipelines for higher resolution and longer clips \cite{ho2022imagen,ho2022video}. Recent systems couple strong VAEs with DiT-style denoisers in latent space -- Stable Video Diffusion \cite{he2022latent} and LTX-Video \cite{hacohen2024ltx} exemplify this trend, targeting practical efficiency (real-time modes) while preserving fidelity. Large-scale open suites (\eg, Wan \cite{wan2025wan} CogVideo \cite{hong2023cogvideo}, CogVideoX\cite{yang2025cogvideox} and HunyuanVideo \cite{kong2024hunyuanvideo}) further push performance via scaled training and improved tokenizers. Our work is orthogonal to the development of visual generation models. Rather than changing the base denoiser or sampler, we analyze how latent frequency characteristics interact with these diffusion-based pipelines, and propose a new training recipe that improves diffusibility across various diffusion and flow matching models.

\vspace{0.1cm}
\noindent\textbf{Frequency Analysis on Neural Networks}\quad With the development of transformer architectures, frequency diagnostics are established to interpret vision transformers \cite{bai2022improving,patro2023scattering,kim2024exploring}. In recent years, a small but growing body of work examines diffusion through a Fourier lens. Falck \etal \cite{falck2025fourier} show that the forward noising disproportionately suppresses high-frequency content, and that reverse processes tend to reconstruct coarse (low-frequency) structure before fine details -- implicating frequency hierarchy as a source of artifacts and inefficiency. Ren \etal \cite{ren2025fds} explicitly inject frequency cues into training or objectives, improving latent editing and related tasks via frequency-aware scores. Closest to our setup, Skorokhodov \etal \cite{skorokhodov2025improving} study spectra of pretrained autoencoders and links latent frequency distributions to degraded diffusion behavior. Distinct from previous studies, our paper focuses on the different responses of the encoder and decoder to input signals in different frequency bands. Our proposed method does not need to re-train or finetune the autoencoder which makes it easier to integrate our method into existing codebases and training recipes.

\section{Frequency Perturbation Analysis}
\label{sec:analysis}

We study the frequency correspondence between RGB and latent spaces in modern autoencoders by perturbing inputs to the encoder/decoder and measuring the induced spectral changes in their outputs. All experiments are conducted on the SOTA deep-compression autoencoder with 128 latent channels (e.g., DC-AE-f32c128) pre-trained on full-band RGB images \cite{chen2025deep}. While prior work has examined frequency characteristics of autoencoder latents \cite{skorokhodov2025improving}, the cross-space correspondence remains unexplored. Our analysis fills this gap and directly informs the design of our method.

\begin{figure}
  \centering
  \includegraphics[width=\linewidth]{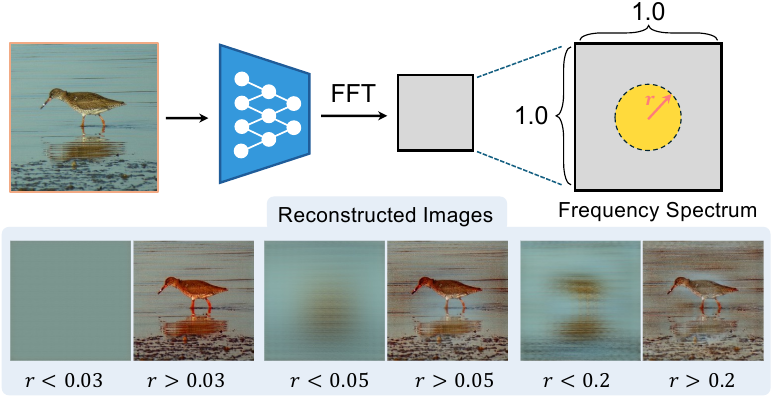}
  \caption{Visualization of images reconstructed from low-frequency and high-frequency embeddings in the latent space. $r$ is the threshold to separate low frequency and high frequency. Please zoom in for more details.}
  \label{fig:decoder_analysis}
\end{figure}

\subsection{Frequency Analysis on Decoder}
\label{sec:decoder_analysis}

First of all, we investigate the contribution of latent embeddings with different frequencies to image reconstruction in the decoding process. Given an image $\bm{X}\in\mathbb{R}^{3\times H\times W}$ in RGB space, we input it into the encoder $\mathcal{E}$ to get the latent embedding $\bm{Z}=\mathcal{E}(\bm{X})\in\mathbb{R}^{C\times H'\times W'}$. We convert the embedding into frequency domain using the 2D fast fourier transform (FFT) on each channel independently, followed by origin shift to move low-frequency component to the center. The resulting frequency profile is 
\begin{equation}\label{eq:fft}\vspace{-0.1cm}
    \bm{Z}_{freq}= \textbf{\texttt{Shift}}(\textbf{\texttt{FFT}}(\bm{Z}))\in\mathbb{C}^{C\times H'\times W'}.
\end{equation}
The value at each location of $\bm{Z}_{freq}$ denotes the amplitude and phase of a specific frequency component. After shifting, the frequency increases from the center to the corner as shown in \cref{fig:decoder_analysis}. We set up a threshold to separate the frequency profile into two parts (frequency higher than threshold and lower than threshold). In practice, we use a circle mask $\bm{M}$ with radius $r$ (as a proxy of frequency threshold) to separate low- and high-frequency signals. The separated frequency profiles are transformed back to the latent space by inverse FFT, which is written as
\begin{equation}\label{eq:low_freq}\vspace{-0.1cm}
    \bm{Z}_{low} = \textbf{\texttt{IFFT}}(\textbf{\texttt{IShift}}(\bm{M} \odot \bm{Z}_{freq})),
\end{equation}
\begin{equation}\label{eq:high_freq}
    \bm{Z}_{high} = \textbf{\texttt{IFFT}}(\textbf{\texttt{IShift}}((1-\bm{M}) \odot \bm{Z}_{freq})),
\end{equation}
where \textbf{\texttt{IFFT}}$(\cdot)$ and \textbf{\texttt{IShift}}$(\cdot)$ are the inverse operations of \textbf{\texttt{FFT}}$(\cdot)$ and \textbf{\texttt{Shift}}$(\cdot)$. $\odot$ is element-wise multiplication.

In \cref{fig:decoder_analysis}, we observe that the RGB images reconstructed from low-frequency latent embeddings are blurry, containing only basic color and layout information. On the contrary, the images reconstructed from high-frequency components include way more details and semantic information. This phenomenon persists when we raise the threshold from 0.05 to 0.20, which reveals the different contributions of different latent frequency bands to image reconstruction.

\finding{2}{\textbf{Finding 1:} Decoder relies more on the information encoded in high-frequency latent embeddings to reconstruct details and semantics in RGB space.}

\subsection{Frequency Analysis on Encoder}
\label{sec:encoder_analysis}

\begin{figure}
  \centering
  \includegraphics[width=\linewidth]{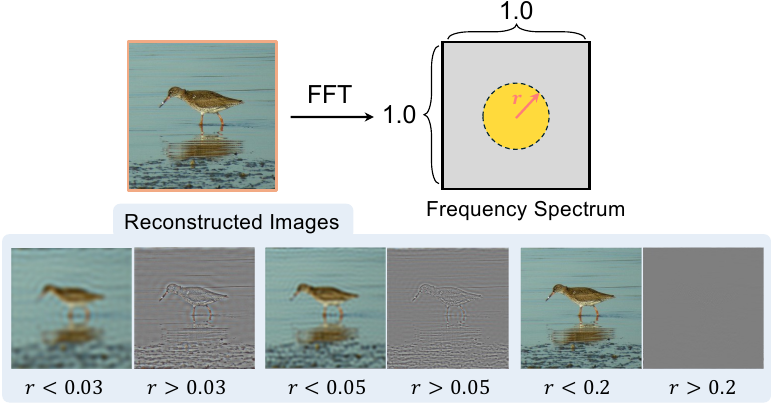}
  \caption{Visualization of images reconstructed from low-frequency and high-frequency components in the RGB space. $r$ is the threshold to separate low frequency and high frequency. Please zoom in for more details.}
  \label{fig:encoder_analysis}
\end{figure}

Based on the analysis in \cref{sec:decoder_analysis} and Finding 1, we conclude that one key factor of achieving a high generation performance is synthesizing reliable high-frequency latent embeddings. To explore the encoding process for different frequencies, we further conduct an analogical analysis on the encoder. Specifically, we transform the RGB image $\bm{X}\in\mathbb{R}^{3\times H\times W}$ to the frequency profile $\bm{X}_{freq}$ by FFT like \cref{eq:fft}. Then we also separate the frequency profile into low- and high-frequency portions using different cutoff thresholds and convert them back to RGB space, akin to \cref{eq:low_freq} and \cref{eq:high_freq}. The visualization is shown in \cref{fig:encoder_analysis}. When the threshold (also represented by the mask radius) is as low as 0.03 or 0.05, we can see a clear separation of information in different frequencies: low frequency encodes colors, layout, size and shapes, while high frequency encodes textures and boundaries. However, when the threshold is raised to 0.20, the image recovered from low-frequency components is almost the same as the original image. The image recovered from high-frequency part contains trivial infromation. Here we come up with the second finding.

\finding{2}{\textbf{Finding 2:} In RGB space, most information of images exist in a narrow low-frequency band, which is different from the distribution in latent space.}

\begin{figure}[t]
  \centering
  \includegraphics[width=\linewidth]{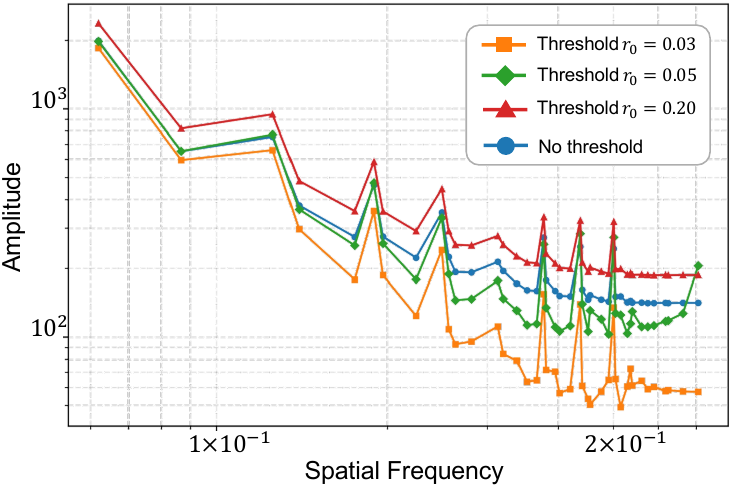}
  \caption{Different frequency distributions in the latent space with regard to the low-pass threshold on RGB images (measured on 50k images). The spatial frequency of x-axis is measured by the distance to the center of frequency spectrum. Both axes are in logarithmic scale. Detailed explanation is described in \cref{sec:analysis}.}
  \label{fig:freq_analysis}
\end{figure}

We further input these images containing only low-frequency signals to a pre-trained encoder to obtain their latent embeddings. We illustrate the corresponding frequency distribution of these latent embeddings in \cref{fig:freq_analysis}. Note that we conduct frequency perturbations in the \textit{RGB} space while measuring the frequency in the \textit{latent} space.

When we use a threshold as low as 0.03, only very low-frequency components are preserved in the input images. We encode the low-frequency signals and obtain the frequency distribution illustrated in the \textcolor{orange}{orange} curve. The energy decreases with the increase of frequency. If the threshold is raised to 0.05 (\textcolor{ForestGreen}{green} curve) to include more high-frequency signal in RGB images, we find a significant improvement in amplitude for high-frequency bands in the latent space while the low frequency remains comparable. We further raise the threshold to 0.20 and obtain the frequency distribution illustrated by the \textcolor{Maroon}{red} curve. A similar change is observed in the latent space -- comparable amplitude in low frequency and increased amplitude in high frequency. Finally, we lift the threshold to include all high frequencies (\textit{i.e.}, original RGB images without threshold). The resultant curve is expected to be above the red curve following the trend in previous experiments. However, we find there is a notable amplitude drop in latent space (shown in the \textcolor{NavyBlue}{blue} curve), especially within high-frequency bands. The result suggests that a portion of very high-frequency RGB signals may prevent the encoding of other high-frequency information. We speculate the reason is that these high-frequency signals trigger aliasing into lower bands (studied in \cite{zhang2019making}) which consumes the capacity for other bands. Combining our observation with Finding 2, we summarize our experiment results in the third finding.

\finding{3}{\textbf{Finding 3:} Extremely high-frequency components in RGB space have marginal contributions to the image quality, but may impede the encoding of other high-frequency signals.}

\section{FreqWarm}

\begin{figure}
  \centering
  \includegraphics[width=\linewidth]{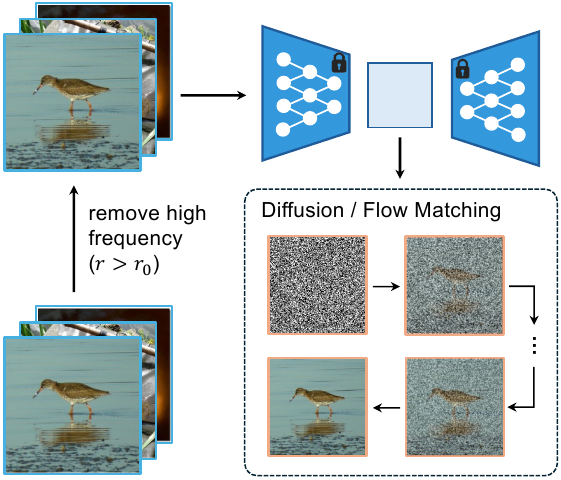}
  \caption{Overview of FreqWarm. We filter out high-frequency components above a frequency threshold $r_0$ in the RGB space. The filtered images are forwarded into a pretrained autoencoder. We train diffusion models or flow matching models on top of the latent space in the early training stage for warm-up. Note that the autoencoder is kept frozen throughout training in our method.}
  \label{fig:method}
  \vspace{-0.2cm}
\end{figure}

Based on the findings in \cref{sec:analysis}, we find a conflict in diffusion model training. When we tokenize the entire images into latent representations, some high-frequency signals block the encoding process, leading to an amplitude drop in high-frequency embeddings (Finding 3). Thus training diffusion models on top of the dominant low-frequency latent embeddings results in a suboptimal performance in fitting high-frequency components. Meanwhile, the decoder heavily relies on the high-frequency latent components to reconstruct images in RGB space (Finding 1). This conflict of encoding and decoding processes becomes more prominent in latent spaces with more dimensionality (shown in \cref{fig:frequency_wrt_channel}). This phenomenon also partially explains the trade-off between reconstruction and generation as shown in \cref{fig:tradeoff}.

To alleviate this conflict, we propose a straightforward and easy-to-implement method, dubbed FreqWarm, to improve the diffusibility in high-dimensional latent spaces defined by autoencoders. As illustrated in \cref{fig:method}, we filter out all the high-frequency signals in RGB images by a threshold $r_0$. Then we tokenize the filtered images using a pre-trained encoder to obtain the latent embeddings with stronger high-frequency components. We warm up diffusion and flow matching model training (from scratch) on top of these embeddings. Then we close the model training with finetuning steps on the full-frequency bands.

\section{Experiments}

\subsection{Dataset and Metrics} 

\textbf{Dataset}\quad We run all the experiments on ImageNet \cite{russakovsky2015imagenet}. Instead of the original ImageNet, we use the face-blurred version to avoid privacy leakage. The only difference is that all faces in the dataset we used are detected and blurred to remove identity. The official training and validation splits are used for diffusion model training. 

\noindent\textbf{Metrics}\quad Following prior work \cite{chen2025dc,chen2025softvq,kouzelis2025eq}, we use generation FID (gFID) \cite{heusel2017gans} and inception score (IS) \cite{salimans2016improved} for quantitative assessment. Since our method does not require re-training existing autoencoders, the reconstruction performance remains the same. We thus do not report the reconstruction quality repeatedly. Given the sensitivity of gFID to the number of images, we synthesize 1000 images for each category (50,000 samples in total) in all experiments for a fair comparison.

\subsection{Implementation Details}

We implement our method on some advanced image/video autoencoders. We directly load the released autoencoder weights for diffusion model training if the checkpoint is publicly released. If the weights are not publicly available, we train the autoencoder using the released code following the same training strategy in the original papers. We follow \cite{chen2025dc} to represent the spatial compression ratio $x$ and channel number $y$ in the form of "f\;$x$\;c\;$y$" (\eg, f32c128 denotes 32$\times$ spatial compression and 128 latent channels).

For diffusion models, we follow the original training settings in their official implementation, except that we increase the batch size to 4096. We use a threshold of $r_0=0.2$ by default in our method if not specified. We train the diffusion model on 32 NVIDIA A100 GPUs for 5-7 days. Specific resources depend on the size and type of the diffusion model.

\begin{table*}[t]
\small
\centering
\setlength{\tabcolsep}{0.18cm}
\begin{tabular}{cccccccc}
\toprule
\multirow{2.5}{*}{\shortstack{Diffusion \\ Model}}  & \multirow{2.5}{*}{Autoencoder}  & \multirow{2.5}{*}{\shortstack{Warm-up \\ in Frequency}}  & \multirow{2.5}{*}{\shortstack{Params \\ (B)}}  & \multicolumn{2}{c}{gFID $\downarrow$} & \multicolumn{2}{c}{IS $\uparrow$} \\
\cmidrule(lr){5-6} \cmidrule(lr){7-8}
& & & & w/o CFG & w/ CFG & w/o CFG & w/ CFG  \\
\midrule
\multirow{7}{*}{DiT-XL \cite{peebles2023scalable}} & \cellcolor[HTML]{FFFFFF} Flux-VAE-f8c16 \cite{labs2025flux1kontextflowmatching} & \cellcolor[HTML]{FFFFFF} None & \cellcolor[HTML]{FFFFFF} 0.67 & \cellcolor[HTML]{FFFFFF} 27.35 & \cellcolor[HTML]{FFFFFF} 8.72 & \cellcolor[HTML]{FFFFFF} 53.09 & \cellcolor[HTML]{FFFFFF} - \\
& \cellcolor[HTML]{FFFFFF} Asym-VAE-f8 \cite{zhu2023designing}  & \cellcolor[HTML]{FFFFFF} None & \cellcolor[HTML]{FFFFFF} 0.67 & \cellcolor[HTML]{FFFFFF} 11.39 & \cellcolor[HTML]{FFFFFF} 2.97 & \cellcolor[HTML]{FFFFFF} - & \cellcolor[HTML]{FFFFFF} - \\
& \cellcolor[HTML]{FFFFFF} SD-VAE-f8c4 \cite{rombach2022high} & \cellcolor[HTML]{FFFFFF} None & \cellcolor[HTML]{FFFFFF} 0.67 & \cellcolor[HTML]{FFFFFF} 12.03 & \cellcolor[HTML]{FFFFFF} 3.04 & \cellcolor[HTML]{FFFFFF} 105.25 & \cellcolor[HTML]{FFFFFF} - \\
\cdashline{2-8}
& \cellcolor[HTML]{FFFFFF} DC-AE-f32c128$^\dagger$ \cite{chen2025deep} & \cellcolor[HTML]{FFFFFF} None  & \cellcolor[HTML]{FFFFFF} 0.67 & \cellcolor[HTML]{FFFFFF} 15.81 & \cellcolor[HTML]{FFFFFF} 3.11 & \cellcolor[HTML]{FFFFFF} 84.41 & \cellcolor[HTML]{FFFFFF} 227.69 \\
& \cellcolor[HTML]{FAEBD7} DC-AE-f32c128$^\dagger$ \cite{chen2025deep}  & \cellcolor[HTML]{FAEBD7} w/ FreqWarm  & \cellcolor[HTML]{FAEBD7} 0.67 & \cellcolor[HTML]{FAEBD7} \textbf{11.02}\diff{-4.79} & \cellcolor[HTML]{FAEBD7} \textbf{2.87}\diff{-0.24} & \cellcolor[HTML]{FAEBD7} \textbf{108.65}\diff{+24.24} & \cellcolor[HTML]{FAEBD7} \textbf{240.05}\diff{+12.36} \\
& \cellcolor[HTML]{FFFFFF} DC-AE-f64c128 \cite{chen2025deep}  & \cellcolor[HTML]{FFFFFF} None  & \cellcolor[HTML]{FFFFFF} 0.67 & \cellcolor[HTML]{FFFFFF} 20.68 & \cellcolor[HTML]{FFFFFF} 5.91 & \cellcolor[HTML]{FFFFFF} 67.69 & \cellcolor[HTML]{FFFFFF} 160.70 \\
& \cellcolor[HTML]{FAEBD7} DC-AE-f64c128 \cite{chen2025deep} & \cellcolor[HTML]{FAEBD7} w/ FreqWarm  & \cellcolor[HTML]{FAEBD7} 0.67 & \cellcolor[HTML]{FAEBD7} 16.63\diff{-4.05} & \cellcolor[HTML]{FAEBD7} 3.74\diff{-2.17} & \cellcolor[HTML]{FAEBD7} 83.71\diff{+16.02} & \cellcolor[HTML]{FAEBD7} 223.05\diff{+62.35} \\
\hline
\multirow{3}{*}{UViT-H \cite{bao2023all}} & \cellcolor[HTML]{FFFFFF} Flux-VAE-f8c16 \cite{labs2025flux1kontextflowmatching} & \cellcolor[HTML]{FFFFFF} None & \cellcolor[HTML]{FFFFFF} 0.50 & \cellcolor[HTML]{FFFFFF} 30.91 & \cellcolor[HTML]{FFFFFF} 12.61 & \cellcolor[HTML]{FFFFFF} - & \cellcolor[HTML]{FFFFFF} - \\
\cdashline{2-8}
& \cellcolor[HTML]{FFFFFF} DC-AE-f64c128 \cite{chen2025deep} & \cellcolor[HTML]{FFFFFF} None & \cellcolor[HTML]{FFFFFF} 0.50 & \cellcolor[HTML]{FFFFFF} 17.34 & \cellcolor[HTML]{FFFFFF} 3.23 & \cellcolor[HTML]{FFFFFF} 84.49 & \cellcolor[HTML]{FFFFFF} 219.30 \\ 
& \cellcolor[HTML]{FAEBD7} DC-AE-f64c128 \cite{chen2025deep} & \cellcolor[HTML]{FAEBD7} w/ FreqWarm & \cellcolor[HTML]{FAEBD7} 0.50 & \cellcolor[HTML]{FAEBD7} \textbf{12.36}\diff{-4.98} & \cellcolor[HTML]{FAEBD7} \textbf{2.76}\diff{-0.47} & \cellcolor[HTML]{FAEBD7} \textbf{108.80}\diff{+24.31} & \cellcolor[HTML]{FAEBD7} \textbf{246.89}\diff{+27.59} \\ 
\hline
\multirow{10}{*}{USiT-H \cite{ma2024sit}} & \cellcolor[HTML]{FFFFFF} Wan2.2-AE-f16c48 \cite{wan2025wan} & \cellcolor[HTML]{FFFFFF} None & \cellcolor[HTML]{FFFFFF} 0.50 & \cellcolor[HTML]{FFFFFF} 43.67 & \cellcolor[HTML]{FFFFFF} 15.13 & \cellcolor[HTML]{FFFFFF} 33.48 & \cellcolor[HTML]{FFFFFF} 88.73 \\
& \cellcolor[HTML]{FAEBD7} Wan2.2-AE-f16c48 \cite{wan2025wan} & \cellcolor[HTML]{FAEBD7} w/ FreqWarm & \cellcolor[HTML]{FAEBD7} 0.50 & \cellcolor[HTML]{FAEBD7} 29.56\diff{-14.11} & \cellcolor[HTML]{FAEBD7} 10.90\diff{-4.23} & \cellcolor[HTML]{FAEBD7} 46.16\diff{+12.68} & \cellcolor[HTML]{FAEBD7} 109.57\diff{+20.84} \\
& \cellcolor[HTML]{FFFFFF} LTX-AE-f32c128 \cite{hacohen2024ltx}  & \cellcolor[HTML]{FFFFFF} None & \cellcolor[HTML]{FFFFFF} 0.50 & \cellcolor[HTML]{FFFFFF} 24.18 & \cellcolor[HTML]{FFFFFF} 6.24 & \cellcolor[HTML]{FFFFFF} 61.60 & \cellcolor[HTML]{FFFFFF} 161.22 \\
& \cellcolor[HTML]{FAEBD7} LTX-AE-f32c128 \cite{hacohen2024ltx}  & \cellcolor[HTML]{FAEBD7} w/ FreqWarm & \cellcolor[HTML]{FAEBD7} 0.50 & \cellcolor[HTML]{FAEBD7} 18.05\diff{-6.13} & \cellcolor[HTML]{FAEBD7} 4.11\diff{-2.13} & \cellcolor[HTML]{FAEBD7} 76.06\diff{+14.46} & \cellcolor[HTML]{FAEBD7} 194.51\diff{+33.29} \\
\cdashline{2-8}
& \cellcolor[HTML]{FFFFFF} DC-AE-f32c128$^\dagger$ \cite{chen2025deep}  & \cellcolor[HTML]{FFFFFF} None & \cellcolor[HTML]{FFFFFF} 0.50 & \cellcolor[HTML]{FFFFFF} 13.84 & \cellcolor[HTML]{FFFFFF} 3.96 & \cellcolor[HTML]{FFFFFF} 85.40 & \cellcolor[HTML]{FFFFFF} 200.70 \\
& \cellcolor[HTML]{FAEBD7} DC-AE-f32c128$^\dagger$ \cite{chen2025deep}  & \cellcolor[HTML]{FAEBD7} w/ FreqWarm  & \cellcolor[HTML]{FAEBD7} 0.50 & \cellcolor[HTML]{FAEBD7} 9.42\diff{-4.42} & \cellcolor[HTML]{FAEBD7} 3.20\diff{-0.76} & \cellcolor[HTML]{FAEBD7} 108.80\diff{+23.40} & \cellcolor[HTML]{FAEBD7} 244.21\diff{+43.51} \\
& \cellcolor[HTML]{FFFFFF} DC-AE-f64c128 \cite{chen2025deep} & \cellcolor[HTML]{FFFFFF} None & \cellcolor[HTML]{FFFFFF} 0.50 & \cellcolor[HTML]{FFFFFF} 9.85 & \cellcolor[HTML]{FFFFFF} 3.13 & \cellcolor[HTML]{FFFFFF} 113.99 & \cellcolor[HTML]{FFFFFF} 207.66 \\
& \cellcolor[HTML]{FAEBD7} DC-AE-f64c128 \cite{chen2025deep} & \cellcolor[HTML]{FAEBD7} w/ FreqWarm & \cellcolor[HTML]{FAEBD7} 0.50 & \cellcolor[HTML]{FAEBD7} \textbf{8.31}\diff{-1.54} & \cellcolor[HTML]{FAEBD7} \textbf{2.64}\diff{-0.49} & \cellcolor[HTML]{FAEBD7} \textbf{130.10}\diff{+16.11} & \cellcolor[HTML]{FAEBD7} \textbf{276.28}\diff{+68.62} \\
& \cellcolor[HTML]{FFFFFF} DC-AE-f128c256 \cite{chen2025deep} & \cellcolor[HTML]{FFFFFF} None & \cellcolor[HTML]{FFFFFF} 0.50 & \cellcolor[HTML]{FFFFFF} 36.71 & \cellcolor[HTML]{FFFFFF} 13.14 & \cellcolor[HTML]{FFFFFF} 49.63 & \cellcolor[HTML]{FFFFFF} 125.33 \\
& \cellcolor[HTML]{FAEBD7} DC-AE-f128c256 \cite{chen2025deep} & \cellcolor[HTML]{FAEBD7} w/ FreqWarm & \cellcolor[HTML]{FAEBD7} 0.50 & \cellcolor[HTML]{FAEBD7} 33.44\diff{-3.27} & \cellcolor[HTML]{FAEBD7} 11.27\diff{-1.87} & \cellcolor[HTML]{FAEBD7} 56.03\diff{+6.40} & \cellcolor[HTML]{FAEBD7} 143.22\diff{+17.89} \\
\hline
\multirow{2.3}{*}{USiT-2B \cite{ma2024sit}} & \cellcolor[HTML]{FFFFFF} DC-AE-f64c128 \cite{chen2025deep} & \cellcolor[HTML]{FFFFFF} None & \cellcolor[HTML]{FFFFFF} 1.58 & \cellcolor[HTML]{FFFFFF} 5.67 & \cellcolor[HTML]{FFFFFF} 3.55 & \cellcolor[HTML]{FFFFFF} 143.73 & \cellcolor[HTML]{FFFFFF} 292.98 \\
& \cellcolor[HTML]{FAEBD7} DC-AE-f64c128 \cite{chen2025deep} & \cellcolor[HTML]{FAEBD7} w/ FreqWarm & \cellcolor[HTML]{FAEBD7} 1.58 & \cellcolor[HTML]{FAEBD7} \textbf{4.77}\diff{-0.90} & \cellcolor[HTML]{FAEBD7} \textbf{3.18}\diff{-0.37} & \cellcolor[HTML]{FAEBD7} \textbf{166.57}\diff{+22.84} & \cellcolor[HTML]{FAEBD7} \textbf{311.79}\diff{+18.81} \\
\bottomrule
\end{tabular}
\begin{tablenotes}
\item $\dagger$ denotes that this autoencoder is reproduced by ourselves using the released codebase.
\end{tablenotes}
\vspace{-0.2cm}
\caption{Results of our method implemented to the latest autoencoders on ImageNet with 512$\times$512 resolution. CFG is short for classifier-free guidance. We set the scale of guidance as 1.5 for all experiments with CFG. The \textcolor{orange}{orange} rows refer to the models trained with the proposed FreqWarm method. The numbers in \textcolor{ForestGreen}{green} are the improvement of our method compared with the baselines without frequency warm-up. Our method outperforms all the counterparts in different combinations of diffusion models and autoencoders.}
\label{tab:sota_cmp}
\vspace{-0.5cm}
\end{table*}

\subsection{Experiments on Latest Autoencoders}

We implement FreqWarm on three high-dimensional autoencoders released recently, including Wan2.2-AE \cite{wan2025wan}, LTX-AE\cite{hacohen2024ltx} and DC-AE \cite{chen2025dc}. Wan2.2-AE and LTX-AE are developed mainly for video tokenization. They can also encode images by setting the number of frames as 1. DC-AE has released the autoencoder weights for 64$\times$ compression and 128 channel (f64c128), but there is no available checkpoints for high-dimensional space with 32$\times$ compression. We train DC-AE-f32c128 on ImageNet from scratch and report the best number we can achieve. To evaluate the generalization of our method, we consider 4 diffusion and flow matching models with different architectures and scales, including DiT-XL \cite{peebles2023scalable}, UViT-H \cite{bao2023all}, USiT-H \cite{ma2024sit} and USiT-2B \cite{ma2024sit}. We also compare with previous low-dimensional autoencoders, such as Flux-VAE \cite{labs2025flux1kontextflowmatching}, Asym-VAE \cite{zhu2023designing} and SD-VAE \cite{rombach2022high}. We do not use classifier-free guidance (CFG) \cite{ho2021classifier} by default unless explicitly stated.

\noindent\textbf{Resolution 512$\times$512}\quad The results are shown in \cref{tab:sota_cmp}. Our method achieves significant improvement on various combinations of high-dimensional autoencoders with different diffusion models, compared with their counterparts trained without frequency warm-up. Specifically, for DiT-XL, our method improves gFID/IS by 4.79/24.24 for DC-AE-f32c128 and 4.05/16.02 for DC-AE-f64c128 without CFG. For USiT-H, our method improves gFID/IS by 14.11/12.68 for Wan2.2-AE, 6.13/14.46 for LTX-AE, 4.42/23.40 for DC-AE-f32c128 and 1.54/16.11 for DC-AE-f64c128. After using CFG in inference, we still observe the prominent gains for all the autoencoders. The results suggest that FreqWarm generalizes well to different autoencoder structures and configurations. With our method, the performance of high-dimensional autoencoders outperform preceding autoencoders with fewer channels. This encouraging results validate that it is feasible to further reduce the number of tokens without performance drop. Our method also leads to reasonable improvement to the larger model -- USiT-2B, which validates the promising scalability of our method to modern large generative models.

\begin{table}[t]
\small
\centering
\begin{tabular}{cccc}
\toprule
Model & Freq. Warm-up  & gFID $\downarrow$  & IS $\uparrow$ \\
\midrule
DiT \cite{peebles2023scalable} & None & 26.30 & 50.36 \\
\rowcolor[HTML]{FAEBD7} DiT \cite{peebles2023scalable} & w/ FreqWarm & \textbf{17.89}\diff{-8.41} & \textbf{72.99}\diff{+22.63} \\
\hline
UViT \cite{bao2023all} & None & 17.99 & 78.15 \\
\rowcolor[HTML]{FAEBD7}UViT \cite{bao2023all}  & w/ FreqWarm & \textbf{12.91}\diff{-5.08} & \textbf{93.50}\diff{+15.35} \\
\hline
USiT \cite{ma2024sit} & None & 15.41 & 81.03 \\
\rowcolor[HTML]{FAEBD7}USiT \cite{ma2024sit} & w/ FreqWarm & \textbf{12.84}\diff{-2.57} & \textbf{94.59}\diff{+13.56} \\
\bottomrule
\end{tabular}
\caption{Experiments on ImageNet with resolution 256$\times$256. All experiments are implemented on top of DC-AE-f32c128 (reproduced by us). The \textcolor{orange}{orange} rows indicate the models trained with our method. The numbers in \textcolor{ForestGreen}{green} are the gains of our method.}
\label{tab:exp_r256}
\vspace{-0.5cm}
\end{table}

\begin{figure*}
  \centering
  \includegraphics[width=\linewidth]{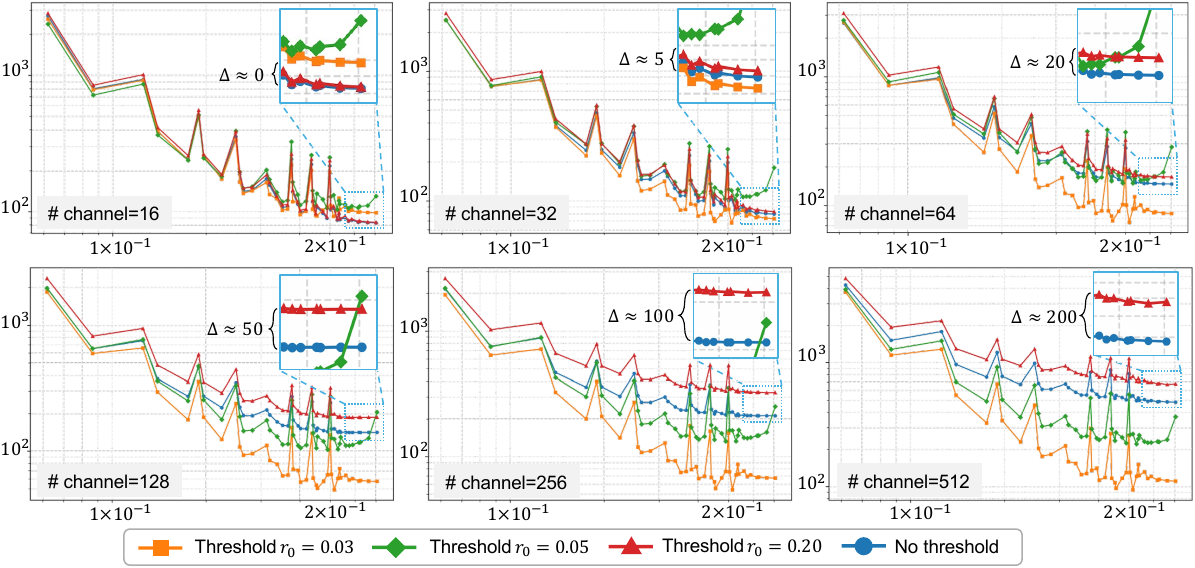}
  \caption{Frequency analysis of latent spaces with channel number ranging from 16 to 512. The x-axis is spatial frequency measured by the distance to the center on frequency spectrum. The y-axis is the amplitude of signals. Note that both axes are in logarithmic scale. The amplitude gap ($\Delta$) of high-frequency bands between threshold $r_0=0.20$ and full band increases with regard to the number of channels.}
  \label{fig:frequency_wrt_channel}
  \vspace{-0.6cm}
\end{figure*}

\noindent\textbf{Resolution 256$\times$256}\quad To validate the generalization of our method, we also run experiments using a lower input resolution -- 256$\times$256, and report the results in \cref{tab:exp_r256}. Our method decreases gFID by 8.41 for DiT, 5.08 for UViT and 2.57 for USiT, showing strong robustness to different resolutions for various diffusion models.

\begin{figure*}
  \centering
  \includegraphics[width=\linewidth]{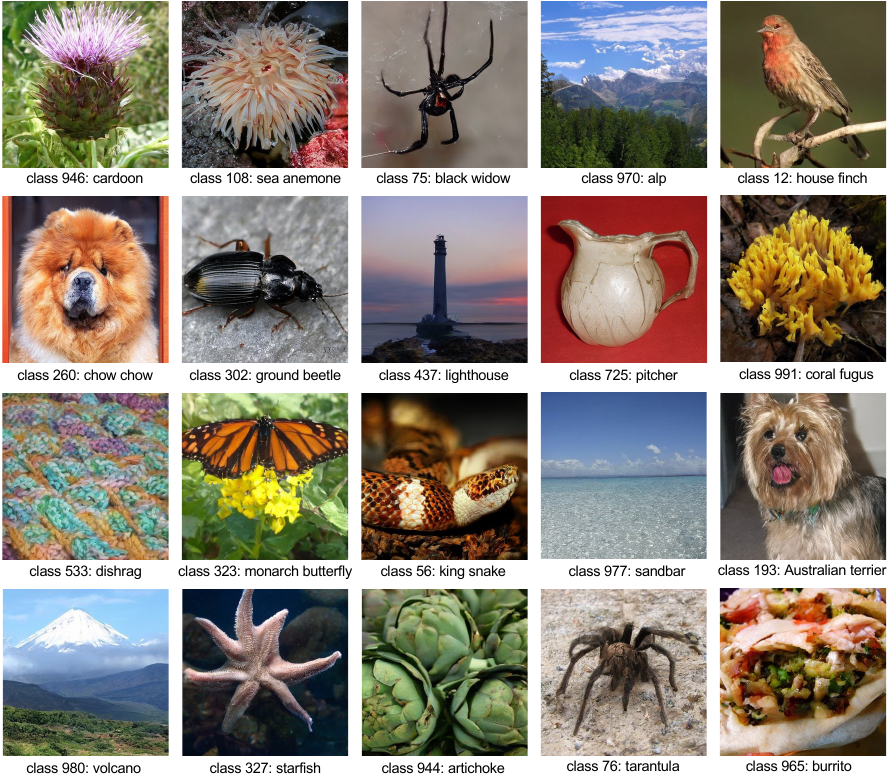}
  \caption{Demonstration of images generated by diffusion models trained with our proposed FreqWarm method.}
  \label{fig:demos}
  \vspace{-0.2cm}
\end{figure*}

\noindent\textbf{Visualization}\quad We illustrate images synthesized by diffusion models trained with our method in \cref{fig:demos}. Our method synthesizes images with high quality across diverse classes.

\begin{table}[t]
\centering
\begin{tabular}{cccc}
\toprule
Configuration  & No Warm-up  & FreqWarm  & \cellcolor[HTML]{FAEBD7} $\Delta$ \\
\midrule
f32c16  & 12.59 & 12.57 & \cellcolor[HTML]{FAEBD7} \textbf{0.02} \\
f32c32  & 5.75 & 5.74 & \cellcolor[HTML]{FAEBD7} \textbf{0.02} \\
f32c64  & 9.97 & 7.20 & \cellcolor[HTML]{FAEBD7} \textbf{2.77} \\
f32c128 & 13.84 & 9.42 & \cellcolor[HTML]{FAEBD7} \textbf{4.42} \\
f32c256 & 42.40 & 33.75 & \cellcolor[HTML]{FAEBD7} \textbf{8.65} \\
f32c512 & 54.84 & 42.66 & \cellcolor[HTML]{FAEBD7} \textbf{12.18} \\
\bottomrule
\end{tabular}
\caption{Analysis of our method implemented to autoencoders with different number of channels. We use gFID ($\downarrow$) as the primary metric to evaluate the performance. The experiments are conducted using different configurations on DC-AE \cite{chen2025deep}.}
\label{tab:ablation_channel_num}
\vspace{-0.4cm}
\end{table}

\subsection{Analysis on Latent Spaces with Different Number of Channels}

To show more insights of our method, we run experiments on DC-AE with different number of channels. The results are demonstrated in \cref{tab:ablation_channel_num}. The channel number varies from 16 to 512 while the compression ratio remains the same. We find our method leads to more gains for latent spaces with more channels. In particular, the performance difference of DC-AE-f32c16 and DC-AE-f32c32 is very marginal due to the low dimension. We think this is a reasonable result, which can be explained by the latent frequency distributions with regard to different channel numbers. Akin to \cref{fig:encoder_analysis}, we do the same analysis on encoders with different channels. As illustrated in \cref{fig:frequency_wrt_channel}, we see the same phenomenon (see Finding 3 in \cref{sec:encoder_analysis}) happens to all high-dimensional encoders: the extremely high-frequency signals hinder the encoding of other components, causing a lower amplitude (the gap between \textcolor{Maroon}{red} curve and \textcolor{NavyBlue}{blue} curve) especially in high-frequency bands. The gap becomes smaller when we reduce the number of channels. We argue that the encoders prioritize low-frequency signals, and then continue to encode high-frequency signals if they have more capacity (\eg, more dimensions). In particular, for f32c16 and f32c32, the two curves almost overlap with each other. The shrinking gap exactly explains the different gains for DC-AE with different dimensions, further supporting our findings and hypothesis. In addition, with our method, we are able to decrease the gFID of DC-AE-f32c512 from 54.84 to 42.66 which is comparable with DC-AE-f32c256 without frequency warm-up (gFID=42.40). Likewise, the performance of DC-AE-f32c128 with FreqWarm (gFID=9.42) is even better than vanilla DC-AE-f32c64 (gFID=9.97).

\begin{table}[t]
\centering
\begin{tabular}{cccc}
\toprule
Model & Threshold $r_0$  & gFID $\downarrow$  & IS $\uparrow$ \\
\midrule
\multirow{4}{*}{\shortstack{DC-AE-f32c128 \cite{chen2025deep} \\ +USiT-H \cite{ma2024sit}}} & 0.05  & 23.11 & 65.50 \\
& \cellcolor[HTML]{FAEBD7} \textbf{0.20} & \cellcolor[HTML]{FAEBD7} \textbf{9.42} & \cellcolor[HTML]{FAEBD7} \textbf{108.80} \\
& 0.40 & 12.88 & 90.49 \\
& 0.60 & 13.24 & 88.71 \\
\bottomrule
\end{tabular}
\caption{Ablation study on the frequency threshold $r_0$. The \textcolor{orange}{orange} row indicates the optimal threshold used in all of our experiments.}
\label{tab:ablation_threshold}
\vspace{-0.3cm}
\end{table}

\subsection{Ablation on Frequency Threshold}

In our method, the key hyperparameter is the frequency threshold $r_0$ used to filter out harmful high-frequency components in RGB space. To test the impact of different thresholds, we use DC-AE-f32c128 and USiT-H as base models and change the threshold from 0.05 to 0.6. Note that we denote the frequency profile as a 1.0$\times$1.0 square, so the range of frequency (radius) is from 0 to $\sqrt{2}\approx0.7$. As demonstrated in \cref{tab:ablation_threshold}, when the threshold is as low as 0.05, only very low-frequency information is preserved, which loses too many details. When the threshold is 0.4 or 0.6, the high-frequency signals are not filtered out thoroughly, leading to suboptimal performance. We find $r_0=0.2$ is an optimal value in our experiments. We also find that removing signals with frequency higher than 0.2 has only trivial impact on image quality while the latent energy is greatly increased (see \cref{fig:frequency_wrt_channel}). This makes 0.2 an ideal trade-off between image quality loss and latent energy drop.

\section{Conclusion}

In this work, we revisit the reconstruction–generation trade-off that emerges when autoencoders define high-dimensional latent spaces for diffusion and flow-matching models. Through a controlled frequency perturbation analysis on both decoder and encoder, we show that decoders strongly depend on high-frequency latent components to recover details, while encoders under-represent exactly these bands once extremely high-frequency RGB signals are present, leading to low latent energy and poor exposure during generative training. Building on these findings, we proposed FreqWarm, a plug-and-play frequency warm-up curriculum that operates entirely on top of frozen, off-the-shelf autoencoders and can be dropped into existing diffusion model training recipes. Extensive experiments on modern high-dimensional tokenizers (Wan2.2-VAE, LTX-VAE, DC-AE) and transformer-based denoisers (DiT, UViT, USiT) demonstrated consistent gains, and the gains increase with the latent dimensionality, confirming that frequency-aware exposure is especially important at scale. Our work points to a co-design of autoencoders and diffusion transformers around explicit frequency budgets, paving the path toward higher compression ratios while preserving comparable generation performance.

{
    \small
    \bibliographystyle{ieeenat_fullname}
    \bibliography{main}

@inproceedings{skorokhodov2025improving,
  title={Improving the Diffusability of Autoencoders},
  author={Skorokhodov, Ivan and Girish, Sharath and Hu, Benran and Menapace, Willi and Li, Yanyu and Abdal, Rameen and Tulyakov, Sergey and Siarohin, Aliaksandr},
  booktitle={Forty-second International Conference on Machine Learning},
  year={2025}
}

@inproceedings{rombach2022high,
  title={High-resolution image synthesis with latent diffusion models},
  author={Rombach, Robin and Blattmann, Andreas and Lorenz, Dominik and Esser, Patrick and Ommer, Bj{\"o}rn},
  booktitle={Proceedings of the IEEE/CVF conference on computer vision and pattern recognition},
  pages={10684--10695},
  year={2022}
}

@inproceedings{peebles2023scalable,
  title={Scalable diffusion models with transformers},
  author={Peebles, William and Xie, Saining},
  booktitle={Proceedings of the IEEE/CVF international conference on computer vision},
  pages={4195--4205},
  year={2023}
}

@inproceedings{chen2025deep,
  title={Deep Compression Autoencoder for Efficient High-Resolution Diffusion Models},
  author={Chen, Junyu and Cai, Han and Chen, Junsong and Xie, Enze and Yang, Shang and Tang, Haotian and Li, Muyang and Han, Song},
  booktitle={The Thirteenth International Conference on Learning Representations},
  year={2025}
}

@misc{labs2025flux1kontextflowmatching,
      title={FLUX.1 Kontext: Flow Matching for In-Context Image Generation and Editing in Latent Space},
      author={Black Forest Labs and Stephen Batifol and Andreas Blattmann and Frederic Boesel and Saksham Consul and Cyril Diagne and Tim Dockhorn and Jack English and Zion English and Patrick Esser and Sumith Kulal and Kyle Lacey and Yam Levi and Cheng Li and Dominik Lorenz and Jonas Müller and Dustin Podell and Robin Rombach and Harry Saini and Axel Sauer and Luke Smith},
      year={2025},
      eprint={2506.15742},
      archivePrefix={arXiv},
      primaryClass={cs.GR},
      url={https://arxiv.org/abs/2506.15742},
}

@article{wan2025wan,
  title={Wan: Open and advanced large-scale video generative models},
  author={Wan, Team and Wang, Ang and Ai, Baole and Wen, Bin and Mao, Chaojie and Xie, Chen-Wei and Chen, Di and Yu, Feiwu and Zhao, Haiming and Yang, Jianxiao and others},
  journal={arXiv preprint arXiv:2503.20314},
  year={2025}
}

@inproceedings{chen2025softvq,
  title={Softvq-vae: Efficient 1-dimensional continuous tokenizer},
  author={Chen, Hao and Wang, Ze and Li, Xiang and Sun, Ximeng and Chen, Fangyi and Liu, Jiang and Wang, Jindong and Raj, Bhiksha and Liu, Zicheng and Barsoum, Emad},
  booktitle={Proceedings of the Computer Vision and Pattern Recognition Conference},
  pages={28358--28370},
  year={2025}
}

@inproceedings{yao2025reconstruction,
  title={Reconstruction vs. generation: Taming optimization dilemma in latent diffusion models},
  author={Yao, Jingfeng and Yang, Bin and Wang, Xinggang},
  booktitle={Proceedings of the Computer Vision and Pattern Recognition Conference},
  pages={15703--15712},
  year={2025}
}

@article{zheng2025diffusion,
  title={Diffusion Transformers with Representation Autoencoders},
  author={Zheng, Boyang and Ma, Nanye and Tong, Shengbang and Xie, Saining},
  journal={arXiv preprint arXiv:2510.11690},
  year={2025}
}

@article{xu2025exploring,
  title={Exploring representation-aligned latent space for better generation},
  author={Xu, Wanghan and Yue, Xiaoyu and Wang, Zidong and Teng, Yao and Zhang, Wenlong and Liu, Xihui and Zhou, Luping and Ouyang, Wanli and Bai, Lei},
  journal={arXiv preprint arXiv:2502.00359},
  year={2025}
}

@article{hacohen2024ltx,
  title={Ltx-video: Realtime video latent diffusion},
  author={HaCohen, Yoav and Chiprut, Nisan and Brazowski, Benny and Shalem, Daniel and Moshe, Dudu and Richardson, Eitan and Levin, Eran and Shiran, Guy and Zabari, Nir and Gordon, Ori and others},
  journal={arXiv preprint arXiv:2501.00103},
  year={2024}
}

@article{zhu2023designing,
  title={Designing a better asymmetric vqgan for stablediffusion},
  author={Zhu, Zixin and Feng, Xuelu and Chen, Dongdong and Bao, Jianmin and Wang, Le and Chen, Yinpeng and Yuan, Lu and Hua, Gang},
  journal={arXiv preprint arXiv:2306.04632},
  year={2023}
}

@article{chen2025hieratok,
  title={Hieratok: Multi-scale visual tokenizer improves image reconstruction and generation},
  author={Chen, Cong and Huang, Ziyuan and Zou, Cheng and Zhu, Muzhi and Ji, Kaixiang and Liu, Jiajia and Chen, Jingdong and Chen, Hao and Shen, Chunhua},
  journal={arXiv preprint arXiv:2509.23736},
  year={2025}
}

@inproceedings{zhang2025diffusion,
  title={Diffusion-4k: Ultra-high-resolution image synthesis with latent diffusion models},
  author={Zhang, Jinjin and Huang, Qiuyu and Liu, Junjie and Guo, Xiefan and Huang, Di},
  booktitle={Proceedings of the Computer Vision and Pattern Recognition Conference},
  pages={23464--23473},
  year={2025}
}

@article{yu2024image,
  title={An image is worth 32 tokens for reconstruction and generation},
  author={Yu, Qihang and Weber, Mark and Deng, Xueqing and Shen, Xiaohui and Cremers, Daniel and Chen, Liang-Chieh},
  journal={Advances in Neural Information Processing Systems},
  volume={37},
  pages={128940--128966},
  year={2024}
}

@article{liu2025hi,
  title={Hi-VAE: Efficient Video Autoencoding with Global and Detailed Motion},
  author={Liu, Huaize and Sun, Wenzhang and Zhang, Qiyuan and Di, Donglin and Gong, Biao and Li, Hao and Wei, Chen and Zou, Changqing},
  journal={arXiv preprint arXiv:2506.07136},
  year={2025}
}

@inproceedings{chen2025dc,
  title={Dc-ae 1.5: Accelerating diffusion model convergence with structured latent space},
  author={Chen, Junyu and Zou, Dongyun and He, Wenkun and Chen, Junsong and Xie, Enze and Han, Song and Cai, Han},
  booktitle={Proceedings of the IEEE/CVF International Conference on Computer Vision},
  pages={19628--19637},
  year={2025}
}

@inproceedings{caron2021emerging,
  title={Emerging properties in self-supervised vision transformers},
  author={Caron, Mathilde and Touvron, Hugo and Misra, Ishan and J{\'e}gou, Herv{\'e} and Mairal, Julien and Bojanowski, Piotr and Joulin, Armand},
  booktitle={Proceedings of the IEEE/CVF international conference on computer vision},
  pages={9650--9660},
  year={2021}
}

@inproceedings{kirillov2023segment,
  title={Segment anything},
  author={Kirillov, Alexander and Mintun, Eric and Ravi, Nikhila and Mao, Hanzi and Rolland, Chloe and Gustafson, Laura and Xiao, Tete and Whitehead, Spencer and Berg, Alexander C and Lo, Wan-Yen and others},
  booktitle={Proceedings of the IEEE/CVF international conference on computer vision},
  pages={4015--4026},
  year={2023}
}

@inproceedings{he2022masked,
  title={Masked autoencoders are scalable vision learners},
  author={He, Kaiming and Chen, Xinlei and Xie, Saining and Li, Yanghao and Doll{\'a}r, Piotr and Girshick, Ross},
  booktitle={Proceedings of the IEEE/CVF conference on computer vision and pattern recognition},
  pages={16000--16009},
  year={2022}
}

@article{zhang2025videorepa,
  title={VideoREPA: Learning Physics for Video Generation through Relational Alignment with Foundation Models},
  author={Zhang, Xiangdong and Liao, Jiaqi and Zhang, Shaofeng and Meng, Fanqing and Wan, Xiangpeng and Yan, Junchi and Cheng, Yu},
  journal={arXiv preprint arXiv:2505.23656},
  year={2025}
}

@inproceedings{zhao2025epsilon,
  title={Epsilon-VAE: Denoising as Visual Decoding},
  author={Zhao, Long and Woo, Sanghyun and Wan, Ziyu and LI, YANDONG and Zhang, Han and Gong, Boqing and Adam, Hartwig and Jia, Xuhui and Liu, Ting},
  booktitle={Forty-second International Conference on Machine Learning},
  year={2025}
}

@article{shi2025latent,
  title={Latent Diffusion Model without Variational Autoencoder},
  author={Shi, Minglei and Wang, Haolin and Zheng, Wenzhao and Yuan, Ziyang and Wu, Xiaoshi and Wang, Xintao and Wan, Pengfei and Zhou, Jie and Lu, Jiwen},
  journal={arXiv preprint arXiv:2510.15301},
  year={2025}
}

@article{bi2025vision,
  title={Vision Foundation Models Can Be Good Tokenizers for Latent Diffusion Models},
  author={Bi, Tianci and Zhang, Xiaoyi and Lu, Yan and Zheng, Nanning},
  journal={arXiv preprint arXiv:2510.18457},
  year={2025}
}

@inproceedings{kouzelis2025eq,
  title={EQ-VAE: Equivariance Regularized Latent Space for Improved Generative Image Modeling},
  author={Kouzelis, Theodoros and Kakogeorgiou, Ioannis and Gidaris, Spyros and Komodakis, Nikos},
  booktitle={Forty-second International Conference on Machine Learning},
  year={2025}
}

@article{zhang2025gpstoken,
  title={GPSToken: Gaussian Parameterized Spatially-adaptive Tokenization for Image Representation and Generation},
  author={Zhang, Zhengqiang and Wu, Rongyuan and Sun, Lingchen and Zhang, Lei},
  journal={arXiv preprint arXiv:2509.01109},
  year={2025}
}

@article{lu2025atoken,
  title={Atoken: A unified tokenizer for vision},
  author={Lu, Jiasen and Song, Liangchen and Xu, Mingze and Ahn, Byeongjoo and Wang, Yanjun and Chen, Chen and Dehghan, Afshin and Yang, Yinfei},
  journal={arXiv preprint arXiv:2509.14476},
  year={2025}
}

@article{sargent2025flow,
  title={Flow to the mode: Mode-seeking diffusion autoencoders for state-of-the-art image tokenization},
  author={Sargent, Kyle and Hsu, Kyle and Johnson, Justin and Fei-Fei, Li and Wu, Jiajun},
  journal={arXiv preprint arXiv:2503.11056},
  year={2025}
}

@inproceedings{lee2025latent,
  title={Latent diffusion models with masked autoencoders},
  author={Lee, Junho and Shin, Jeongwoo and Choi, Hyungwook and Lee, Joonseok},
  booktitle={Proceedings of the IEEE/CVF International Conference on Computer Vision},
  pages={17422--17431},
  year={2025}
}

@article{qiu2025image,
  title={Image Tokenizer Needs Post-Training},
  author={Qiu, Kai and Li, Xiang and Chen, Hao and Kuen, Jason and Xu, Xiaohao and Gu, Jiuxiang and Luo, Yinyi and Raj, Bhiksha and Lin, Zhe and Savvides, Marios},
  journal={arXiv preprint arXiv:2509.12474},
  year={2025}
}

@article{vallaeys2025ssdd,
  title={SSDD: Single-Step Diffusion Decoder for Efficient Image Tokenization},
  author={Vallaeys, Th{\'e}ophane and Verbeek, Jakob and Cord, Matthieu},
  journal={arXiv preprint arXiv:2510.04961},
  year={2025}
}

@article{wu2025h3ae,
  title={H3AE: High Compression, High Speed, and High Quality AutoEncoder for Video Diffusion Models},
  author={Wu, Yushu and Li, Yanyu and Skorokhodov, Ivan and Kag, Anil and Menapace, Willi and Girish, Sharath and Siarohin, Aliaksandr and Wang, Yanzhi and Tulyakov, Sergey},
  journal={arXiv preprint arXiv:2504.10567},
  year={2025}
}

@article{medi2025missing,
  title={Missing Fine Details in Images: Last Seen in High Frequencies},
  author={Medi, Tejaswini and Wang, Hsien-Yi and Rampini, Arianna and Keuper, Margret},
  journal={arXiv preprint arXiv:2509.05441},
  year={2025}
}

@inproceedings{mahapatra2025progressive,
  title={Progressive Growing of Video Tokenizers for Temporally Compact Latent Spaces},
  author={Mahapatra, Aniruddha and Mai, Long and Bourgin, David and Zhang, Yitian and Liu, Feng},
  booktitle={Proceedings of the IEEE/CVF International Conference on Computer Vision},
  pages={17629--17639},
  year={2025}
}

@article{he2022latent,
  title={Latent video diffusion models for high-fidelity long video generation},
  author={He, Yingqing and Yang, Tianyu and Zhang, Yong and Shan, Ying and Chen, Qifeng},
  journal={arXiv preprint arXiv:2211.13221},
  year={2022}
}

@inproceedings{bao2023all,
  title={All are worth words: A vit backbone for diffusion models},
  author={Bao, Fan and Nie, Shen and Xue, Kaiwen and Cao, Yue and Li, Chongxuan and Su, Hang and Zhu, Jun},
  booktitle={Proceedings of the IEEE/CVF conference on computer vision and pattern recognition},
  pages={22669--22679},
  year={2023}
}

@inproceedings{ma2024sit,
  title={Sit: Exploring flow and diffusion-based generative models with scalable interpolant transformers},
  author={Ma, Nanye and Goldstein, Mark and Albergo, Michael S and Boffi, Nicholas M and Vanden-Eijnden, Eric and Xie, Saining},
  booktitle={European Conference on Computer Vision},
  pages={23--40},
  year={2024},
  organization={Springer}
}

@article{ho2022imagen,
  title={Imagen video: High definition video generation with diffusion models},
  author={Ho, Jonathan and Chan, William and Saharia, Chitwan and Whang, Jay and Gao, Ruiqi and Gritsenko, Alexey and Kingma, Diederik P and Poole, Ben and Norouzi, Mohammad and Fleet, David J and others},
  journal={arXiv preprint arXiv:2210.02303},
  year={2022}
}

@article{ho2022video,
  title={Video diffusion models},
  author={Ho, Jonathan and Salimans, Tim and Gritsenko, Alexey and Chan, William and Norouzi, Mohammad and Fleet, David J},
  journal={Advances in neural information processing systems},
  volume={35},
  pages={8633--8646},
  year={2022}
}

@inproceedings{hong2023cogvideo,
  title={CogVideo: Large-scale Pretraining for Text-to-Video Generation via Transformers},
  author={Hong, Wenyi and Ding, Ming and Zheng, Wendi and Liu, Xinghan and Tang, Jie},
  booktitle={The Eleventh International Conference on Learning Representations},
  year={2023}
}

@inproceedings{yang2025cogvideox,
  title={CogVideoX: Text-to-Video Diffusion Models with An Expert Transformer},
  author={Yang, Zhuoyi and Teng, Jiayan and Zheng, Wendi and Ding, Ming and Huang, Shiyu and Xu, Jiazheng and Yang, Yuanming and Hong, Wenyi and Zhang, Xiaohan and Feng, Guanyu and others},
  booktitle={The Thirteenth International Conference on Learning Representations},
  year={2025}
}

@article{kong2024hunyuanvideo,
  title={Hunyuanvideo: A systematic framework for large video generative models, 2025},
  author={Kong, W and Tian, Q and Zhang, Z and Min, R and Dai, Z and Zhou, J and Xiong, J and Li, X and Wu, B and Zhang, J and others},
  journal={URL https://arxiv. org/abs/2412.03603},
  year={2024}
}

@article{falck2025fourier,
  title={A Fourier Space Perspective on Diffusion Models},
  author={Falck, Fabian and Pandeva, Teodora and Zahirnia, Kiarash and Lawrence, Rachel and Turner, Richard and Meeds, Edward and Zazo, Javier and Karmalkar, Sushrut},
  journal={arXiv preprint arXiv:2505.11278},
  year={2025}
}

@inproceedings{ren2025fds,
  title={FDS: Frequency-Aware Denoising Score for Text-Guided Latent Diffusion Image Editing},
  author={Ren, Yufan and Jiang, Zicong and Zhang, Tong and Forchhammer, S{\o}ren and S{\"u}sstrunk, Sabine},
  booktitle={Proceedings of the Computer Vision and Pattern Recognition Conference},
  pages={2651--2660},
  year={2025}
}

@inproceedings{bai2022improving,
  title={Improving vision transformers by revisiting high-frequency components},
  author={Bai, Jiawang and Yuan, Li and Xia, Shu-Tao and Yan, Shuicheng and Li, Zhifeng and Liu, Wei},
  booktitle={European Conference on Computer Vision},
  pages={1--18},
  year={2022},
  organization={Springer}
}

@inproceedings{kim2024exploring,
  title={Exploring adversarial robustness of vision transformers in the spectral perspective},
  author={Kim, Gihyun and Kim, Juyeop and Lee, Jong-Seok},
  booktitle={Proceedings of the IEEE/CVF Winter Conference on Applications of Computer Vision},
  pages={3976--3985},
  year={2024}
}

@article{patro2023scattering,
  title={Scattering vision transformer: Spectral mixing matters},
  author={Patro, Badri and Agneeswaran, Vijay},
  journal={Advances in Neural Information Processing Systems},
  volume={36},
  pages={54152--54166},
  year={2023}
}

@article{russakovsky2015imagenet,
  title={Imagenet large scale visual recognition challenge},
  author={Russakovsky, Olga and Deng, Jia and Su, Hao and Krause, Jonathan and Satheesh, Sanjeev and Ma, Sean and Huang, Zhiheng and Karpathy, Andrej and Khosla, Aditya and Bernstein, Michael and others},
  journal={International journal of computer vision},
  volume={115},
  number={3},
  pages={211--252},
  year={2015},
  publisher={Springer}
}

@inproceedings{ho2021classifier,
  title={Classifier-Free Diffusion Guidance},
  author={Ho, Jonathan and Salimans, Tim},
  booktitle={NeurIPS 2021 Workshop on Deep Generative Models and Downstream Applications},
  year={2021}
}

@article{heusel2017gans,
  title={Gans trained by a two time-scale update rule converge to a local nash equilibrium},
  author={Heusel, Martin and Ramsauer, Hubert and Unterthiner, Thomas and Nessler, Bernhard and Hochreiter, Sepp},
  journal={Advances in neural information processing systems},
  volume={30},
  year={2017}
}

@article{salimans2016improved,
  title={Improved techniques for training gans},
  author={Salimans, Tim and Goodfellow, Ian and Zaremba, Wojciech and Cheung, Vicki and Radford, Alec and Chen, Xi},
  journal={Advances in neural information processing systems},
  volume={29},
  year={2016}
}

@inproceedings{zhang2019making,
  title={Making convolutional networks shift-invariant again},
  author={Zhang, Richard},
  booktitle={International conference on machine learning},
  pages={7324--7334},
  year={2019},
  organization={PMLR}
}
}

\end{document}